\documentclass[letterpaper]{article} 
\usepackage{aaai23}  
\usepackage{times}  
\usepackage{helvet}  
\usepackage{courier}  
\usepackage[hyphens]{url}  
\usepackage{graphicx} 
\usepackage{natbib}  
\usepackage{caption} 
\usepackage{algorithm}
\usepackage{algorithmic}

\usepackage{color}

\newcommand{\ie}{\textit{i}.\textit{e}. }

\newtheorem{definition}{Definition}

\newtheorem{claim}{Claim}

\newcommand{\E}{\mathbb{E}}

\usepackage{bm,amsmath,amsfonts}

\def\vmu{{\bm{\mu}}}

\def\vsigma{{\bm{\sigma}}}

\def\vu{{\bm{u}}}
\def\vv{{\bm{v}}}

\def\vx{{\bm{x}}}
\def\vy{{\bm{y}}}

\def\sF{{\mathbb{F}}}

\def\sN{{\mathbb{N}}}

\def\sR{{\mathbb{R}}}

\usepackage{newfloat}
\usepackage{listings}
\DeclareCaptionStyle{ruled}{labelfont=normalfont,labelsep=colon,strut=off} 
\floatstyle{ruled}
\newfloat{listing}{tb}{lst}{}
\floatname{listing}{Listing}
\title{Online Symbolic Regression with Informative Query}
\author {
    Pengwei Jin \textsuperscript{\rm 1,\rm 2,\rm 3},
    Di Huang \textsuperscript{\rm 1,\rm 2,\rm 3},
    Rui Zhang \textsuperscript{\rm 1, \rm 3},
    Xing Hu \textsuperscript{\rm 1},
    Ziyuan Nan \textsuperscript{\rm 1,\rm 2,\rm 3},
    Zidong Du \textsuperscript{\rm 1},

    Qi Guo \textsuperscript{\rm 1},
    Yunji Chen \textsuperscript{\rm 1,\rm 2}\thanks{Yunji Chen (cyj@ict.ac.cn) is the corresponding author.}
}
\affiliations {
    \textsuperscript{\rm 1} State Key Lab of Processors, Institute of Computing Technology, CAS
\\
    \textsuperscript{\rm 2} University of Chinese Academy of Sciences\\
    \textsuperscript{\rm 3} Cambricon Technologies\\
    \{jinpengwei20z, huangdi20b, zhangrui, huxing, nanziyuan21s, duzidong, guoqi, cyj\}@ict.ac.cn
}

\usepackage{bibentry}

\begin{document}

\maketitle

\begin{abstract}
Symbolic regression, the task of extracting mathematical expressions from the observed data $\{ \vx_i, y_i \}$,
plays a crucial role in scientific discovery.
Despite the promising performance of existing methods, most of them conduct symbolic regression in an \textit{offline} setting.
That is, they treat the observed data points as given ones that are simply sampled from uniform distributions without exploring the expressive potential of data.
However, for real-world scientific problems, the data used for symbolic regression are usually actively obtained by doing experiments, which is an \textit{online} setting.
Thus, how to obtain informative data that can facilitate the symbolic regression process is an important problem that remains challenging. 

In this paper, we propose QUOSR, a \textbf{qu}ery-based framework for \textbf{o}nline \textbf{s}ymbolic \textbf{r}egression that can automatically obtain informative data in an iterative manner.
Specifically, at each step, QUOSR receives historical data points, generates new $\vx$, and then queries the symbolic expression to get the corresponding $y$, where the $(\vx, y)$ serves as new data points.
This process repeats until the maximum number of query steps is reached.
To make the generated data points informative, we implement the framework with a neural network and train it by maximizing the mutual information between generated data points and the target expression.
Through comprehensive experiments, we show that QUOSR can facilitate modern symbolic regression methods by generating informative data. 
\end{abstract}

\section{Introduction}
\label{sec:Introduction}

Symbolic regression plays a central role in scientific discovery, which extracts the underlying mathematical relationship between variables from observed data. 
Formally, given a physical system $f$ and $N$ data points $D=\{ (\vx_i, y_i) \}_{i=1}^N $, where $\vx_i \in \mathbb{R}^m$ and $y_i=f(\vx_i) \in \mathbb{R}$,
symbolic regression tries to find a mathematical expression $\hat{f}:\mathbb{R}^m \rightarrow \mathbb{R}$ that best fits the physical system $f$.

Conventional symbolic regression methods include genetic evolution algorithms~\cite{augusto2000symbolic, schmidt2009distilling, chen2017elite}, and neural network methods~\cite{udrescu2020ai, Petersen2021DeepSR, Valipour2021SymbolicGPTAG}.
Despite their promising performance, most of them perform symbolic regression in an \textit{offline} setting.
That is, they take the data points $D=\{ (\vx_i, y_i) \}_{i=1}^N $ as given ones and simply sample them from uniform distributions without exploring the expressive potential of data.
In real-world scientific discovery, however, data can be collected actively by doing specific experiments~\cite{hernandez2019fast, weng2020simple}, which is an \textit{online} setting.
Overall, online symbolic regression is a worth-studying and challenging problem that underexplored.

The unique problem introduced by online symbolic regression is how to actively collect informative data that best distinguish the target expression from other similar expressions. Specifically, this process aims to actively collect data by generating informative input data $\vx$ (\ie queries) and then querying $f$ to get the output data $y=f(\vx)$. Considering the regression process to use data $D=\{ (\vx_i, y_i) \}_{i=1}^N $ for finding the mathematical expression $\hat{f}$ that best fits the physical system can be borrowed directly from offline symbolic regression, we mainly focus on the key problem of how to get the informative data for symbolic regression. 

In this paper, we propose QUOSR, a \textbf{qu}ery-based framework that can acquire informative data for \textbf{o}nline \textbf{s}ymbolic \textbf{r}egression.
QUOSR works in an iterative manner: at each query step, it receives historical data points, generates new $\vx$, and then queries the target $f$ to get the corresponding $y$, where the $(\vx, y)$ serves as new data points.
This process repeats until the maximum number of query steps is reached.
Our contribution is twofold: giving theoretical guidance on the learning of the query process and implementing QUOSR for online symbolic regression.

Specifically, we theoretically show the relationship between informative data and mutual information, which can guide the generation of queries.
This relationship can be proved by modeling the query process as the expansion of a decision tree, where the nodes
correspond to the query selections $\vx$, the edges correspond to the responses $y$, and the leaves correspond to the physical systems $f$
Although the computation of mutual information is intractable, we can instead optimize QUOSR by minimizing the InfoNCE loss in contrastive learning~\cite{Oord2018RepresentationLW} which is the lower bound of mutual information.

However, it is still difficult to practically implement QUOSR for online symbolic regression for two reasons.
(1) 
The amount of information contained in each data point is so small that, to obtain enough information, the number of query steps becomes extremely large, which makes the query process not only time-consuming but also hard to optimize.
(2)
The InfoNCE loss needs to calculate the similarity between queries and the target system $f$ in a latent space, but $f$ is hard to represent.
Conventionally, $f$ would be represented by expression strings, which is sub-optimal.
First, the same function $f$ can be expressed by different symbolic forms.
For example, $f(x)=cos(x)$ can also be written as $f(x)=cos(x+2k\pi)$ or $f(x)=sin(x+\frac{4k+1}{2}\pi)$, where $k \in \sN$, with exactly the same functionality.
Second, a small change to the expression can make a huge difference in functionality.

To tackle these challenges, we propose the query-by-distribution strategy and the modified InfoNCE loss.
First, to increase the amount of information obtained in each query step, the query-by-distribution strategy aims to find informative function features, which can be represented by a set of data instead of a single data point.
Second, to eliminate the influence of the expression representation problem, we remove the requirement of the calculation of similarity between data and expressions in the InfoNCE loss. Instead, we show that mutual information between data and expressions can be estimated by simply measuring the similarity among different sets of queries.
 
Through comprehensive experiments, QUOSR shows its advantage in exploring informative data for online symbolic regression. We combine QUOSR with a transformer-based method called SymbolicGPT~\cite{Valipour2021SymbolicGPTAG} and find an over 11\% improvement in average $R^2$.

\section{Related Work} 

\paragraph{Symbolic regression.}
Symbolic regression in an offline setting has been studied for years~\cite{koza1994genetic, martius2016extrapolation, sahoo2018learning,lample2019deep, la2021contemporary, xing2021automated, zheng2022symbolic}.
Traditionally, methods based on genetic evolution algorithms have been utilized to tackle offline symbolic regression~\cite{augusto2000symbolic, schmidt2009distilling, arnaldo2014multiple, la2018learning, virgolin2021improving, mundhenk2021symbolic}.
Many recent methods leverage neural networks with the development of deep learning.
AI Feynman~\cite{udrescu2020ai} recursively decomposes an expression with physics-inspired techniques and trains a neural network to fit complex parts.
\citet{Petersen2021DeepSR} proposes a reinforcement learning based method, where expressions are generated by an LSTM.
Recently, some transformer-based methods are proposed~\cite{Biggio2021NeuralSR, Kamienny2022EndtoendSR}.
SymbolicGPT~\cite{Valipour2021SymbolicGPTAG} generates a large amount of expressions and pretrains a transformer-based model to predict expressions.
Similar to QUOSR, \citet{ansari2022iterative} and \citet{haut2022active} also get data points from a physical system. However, there are still two main differences. For purpose, they focus on training efficiency which is an active learning style, while we focus both on the training and inference. For method, theirs are strongly correlated with the SR process, while QUOSR is a general framework since its query process is decomposed from SR.

\paragraph{Active learning.}
Active learning is a subfield of machine learning in which learners ask the oracle to label some representative examples~\cite{settles2009active}. 
Three main scenarios in which learners can ask queries are considered in active learning:
(1) pool-based sampling~\cite{lewis1994sequential} which evaluates all examples in the dataset and samples some of them,
(2) stream-based selective sampling~\cite{cohn1994improving, angluin2001queries} in which each example in the dataset is independently assessed for the amount of information,
and (3) membership query synthesis~\cite{angluin1988queries, king2004functional} in which examples for query can be generated by learners.
Notice that active learning assumes only one oracle to give labels thus learners just need to model one, but each expression is an oracle in QUOSR which makes selecting examples much more difficult.
Also, in active learning, examples are only sampled in the training process to make use of abundant unlabeled data but QUOSR aims to find informative examples both in the training and inference process to assist the following expression generation stage.

\paragraph{Learning to acquire information.}
\citet{pu2017learning} and \citet{pu2018selecting} study the query problem in a finite space aiming to select representative ones from a set of examples. 
\citet{Huang2022NeuralPS} propose an iterative query-based framework in program synthesis and generalize the query problem to a nearly infinite space.
They take the symbolic representation of programs as reference substances during optimization and adopt a query-by-point strategy, where the framework iterates for several steps and only one data point is generated at each step, which fails on symbolic regression.

\section{Problem statement}
In this section, we make a formal statement for online symbolic regression to make our problem clear.
\begin{definition}[The physical system $f$]
\label{def:f}
The physical system $f$ is a function $f:\sR^M\rightarrow\sR$.
Given an input data $\vx\in\sR^M$, $f$ can respond with $y=f(\vx)$.
\end{definition}
Intuitively, $f$ can be seen as scientific rules, and the input data $\vx$ and the response $y$ can be seen as experimental results.
\begin{definition}[Online symbolic regression]
\label{def:goal of osr}
Given a physical system $f \in \sF$, where $\sF$ denotes the set of all possible systems, the online symbolic regression aims to find a mathematical expression $\hat{f}$ that best fits $f$ by interacting with $f$, \ie $\forall \vx\in\sR^M, |\hat{f(\vx)}-f(\vx)|<\tau$, where $\tau$ is the error tolerance.
\end{definition}
Unlike conventional symbolic regression tasks, online symbolic regression allows interactions with the physical system $f$, which is a more common setting in scientific discovery.

\section{Methods}
\label{sec:methods}
In this section, first, we make a detailed description of the query-based framework, including the working process of the framework, what is "informative", and how to obtain informative data theoretically.
Then, we point out the difficulties of applying the query-based framework and give out corresponding solutions.
At the end of this section, we present the architecture details and the training algorithm of the query-based framework.

\begin{algorithm}[t]
    \caption{The query-based framework}
    \label{alg:online symbolic regression}
    Query
    \begin{algorithmic}[1]
            \STATE physical system $f$, max query steps $K$, query module $Q$, initial data points $D$.
            \FOR{$ k \in \{1 \ldots K \}$}
                \STATE $\vx \gets Q(D)$
                \STATE $y \gets f(\vx)$
                \STATE $D \gets D \cup \{(\vx, y)\}$
            \ENDFOR \\
            \RETURN $D$
    \end{algorithmic}
    Regression
    \begin{algorithmic}[1]
            \STATE data points $D$, regression algorithm $R$
            \STATE $\hat{f} \gets R(D)$
            \RETURN $\hat{f}$
    \end{algorithmic}
\end{algorithm}

\subsection{The Query-based Framework}
We propose the query-based framework which aims to generate informative data by querying the target physical system.
The query-based framework works in an iterative manner that it queries the target system for several steps with a data point generated at each step.
This process is shown in Algorithm~\ref{alg:online symbolic regression}.
Specifically, at each step, the query-based framework receives the historical data points $D$ and generates the next query $\vx$ with a trained query module $Q$ according to $D$.
Then, $\vx$ is used to query the target system $f$ and get the corresponding output $y=f(\vx)$.
Finally, the newly queried data point $(\vx, y)$ is added to $D$, and the next query step begins.
This process repeats until the maximum number of query steps is reached.
After collecting the data points $D$, the regression process begins to fit the symbolic expression $\hat{f}$, which can be done by conventional symbolic regression algorithms.

Obviously, the design of the query module $Q$ is the central problem of the query process.
To make the design principle clear, two questions need to be answered.
That is, what is informative data, and how can we get it?

What is informative data?
Intuitively, if data $D_B$ can lead us to the target $f$ but $D_A$ cannot, then we would say $D_B$ is more informative than $D_A$.
Further on, if data $D_C$ can also lead us to the target $f$ and it contains fewer data points than $D_B$, then we would say $D_C$ is even more informative than $D_B$.
Based on this observation, we can conclude that informative means to find the target $f$ with as few data points as possible. 
Formally, we give the following definitions.
\begin{definition}[Distinguish]
\label{def:distinguishable}
Given the target physical system $f\in \sF$, and a set of input data $Q = \{\vx | \vx \in \sR^m \}$, we say that $f$ is distinguished from $\sF$ by Q if and only if $\forall f_i \in \sF, \exists \vx \in Q, f(\vx) \neq f_i(\vx)$.
\end{definition}
\begin{definition}[Informative queries]
\label{def:informative data}
Given the target physical system $f\in \sF$, and two sets of queries $Q_i$ and $Q_j$,
we say $Q_i$ is more informative than $Q_j$ if and only if $Q_i$ can distinguish $f$ from $\sF$ but $Q_j$ cannot, or they can both distinguish $f$ from $\sF$ but $|Q_i| < |Q_j|$.
\end{definition}

How can we obtain informative data?
From the perspective of the decision process, we can model the query process as the expansion of a decision tree.
Specifically, the nodes correspond to the query selections $\vx$, the edges correspond to the responses $y$, and the leaves correspond to the physical systems $f$.
Then, the problem of obtaining informative data can be reinterpreted as the minimization of the average length of decision paths.
\begin{definition}[Average decision path]
\label{def:average decision path}
Given a decision tree with a leaf denoted as $f\in\sF$, the decision path can be denoted as $L(f)$, and the probability of arriving at $f$ can be denoted as $P(f)$. Then, the average decision path can be defined as $\bar{L}=\sum_f P(f)L(f)$.
\end{definition}

Obviously, the average decision path here is equivalent to the average number of query steps.
Traditionally, the problem of finding the average decision path can be solved by maximizing the information gain in decision node expansion, which is the same as maximizing mutual information~\cite{Gallager1968InformationTA}.
Thus, with the help of decision trees, we conclude that informative data is the data that can maximize the mutual information with the target functions.
Formally, we give such claim:
\begin{claim}[Mutual information guides the query]
The collection of informative data can be guided by mutual information:
\begin{equation}
\label{eq:mi}
\begin{aligned}
D^* 
&= \mathop{\arg\max}\limits_{D} I(F;D).
\end{aligned}
\end{equation}
\end{claim}
Here $F$ denotes the variable that represents functions.
The detailed proof is shown in Appendix A.

\begin{figure*}[ht]
    \centering
    \includegraphics[scale=0.41]{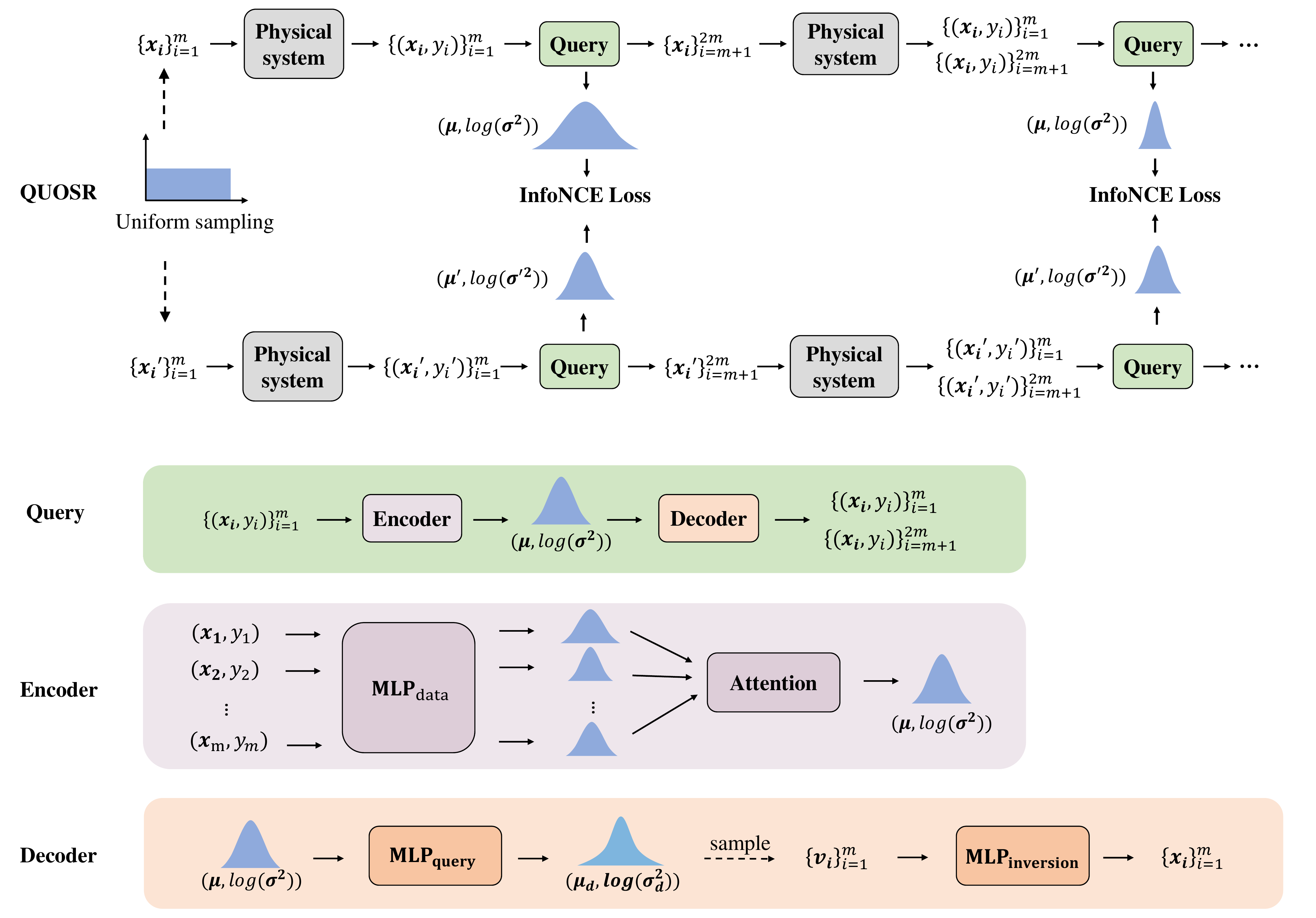}
    \caption{Overview of QUOSR.}
    \label{fig:overview}
\end{figure*}

\subsection{The Modified InfoNCE}
\label{sec:the modified infonce}

\subsubsection{InfoNCE}
Although Equation~\ref{eq:mi} has shown the optimization direction of the query process, $I(F;D)$ is impossible to compute due to the large space of data and expressions.
Fortunately, recent works in contrastive learning show that the InfoNCE loss estimates the lower bound of mutual information~\cite{Oord2018RepresentationLW,PooleOOAT19}.
\begin{equation}
\label{eq:infonce}
L_{NCE} = -\E[log(\frac{exp(sim(d_i, f_i))}{\sum_{j=1}^N exp(sim(d_i, f_j))})],
\end{equation}
and
\begin{equation}
I(F;D) \geq log(B)-L_{NCE}
\end{equation} 
where $d$ denotes the data points, $f$ denotes the physical system, $B$ denotes the batch size, $i,j$ denotes the sample index, and $sim$ denotes a similarity function.
Thus, we can implement the query process as a query network, and then optimize it with InfoNCE to obtain informative queries.
Specifically, in contrastive learning, we can construct several positive and negative pairs, and the InfoNCE loss guides the network to maximize the similarity of positive pairs and minimize the similarity of negative pairs.
For the query network, we can construct positive pairs as queried data points and corresponding physical system $(d_i, f_i)$ and negative pairs as others $(d_i, f_{j \neq i})$.

\subsubsection{The Modified InfoNCE}
According to Equation~\ref{eq:infonce}, we can simply design the query network with a data encoder to extract the embedding of $d_i$ and an expression encoder to extract the embedding of $f_i$, and then compute their similarity to get the InfoNCE loss.
However, the expression encoder is hard to optimize due to several reasons.
First, the same function $f$ can be expressed by different symbolic forms, which improves the difficulty of learning.
For example, $f(x)=cos(x)$ can also be written as $f(x)=cos(x+2k\pi)$ or $f(x)=sin(x+\frac{4k+1}{2}\pi)$, where $k \in \sN$, with exactly the same functionality.
Second, a small change to the expression can make a huge difference in functionality, like the negative sign $-$.

To solve this problem, we propose a modified InfoNCE loss that guides the learning process without the requirement for a good expression encoder:
\begin{equation}
\label{eq:modified infonce}
\begin{aligned}
L'_{NCE} =& -\E_{(d_i, d'_i)\sim\E[P_{d|f}P_{d'|f}]} \\
& [log(\frac{exp(sim(d_i, d'_i))}{\frac{1}{N}\sum_{j=1}^N exp(sim(d_i, d'_j))})],
\end{aligned}
\end{equation}
where $d'$ denotes another set of queried data points with different initial conditions from $d$.
Specifically, we first sample physical system $f$ from space $\sF$, sample two different groups of data points $d_i$ and $d'_i$ from a Uniform distribution as the initial condition and then get two different data points by querying $f$ and modifying $d_i$ and $d'_i$ (adding the corresponding data point to $d_i$ and another to $d'_i$), and this process repeats for $N$ times to get $N$ different $(d_i, d'_i)$ pairs to calculate the loss.

Intuitively, the main difference between Equation~\ref{eq:modified infonce} and Equation~\ref{eq:infonce} is that the former replaces the expression $f$ with another set of queried points $d'$, which avoids the expression representation problems mentioned above.
In practice, we share parameters between the data encoders for $d$ and $d'$, which is the classical siamese architecture used in contrastive learning~\cite{Chen2020ASF}.

We prove that optimizing $L'_{NCE}$ can also achieve the purpose of maximizing mutual information $I(F;D)$:
\begin{claim}[$L'_{NCE}$ bounds mutual information]
\begin{equation}
\label{eq:min}
\begin{aligned}
-L'_{NCE} &\leq KL(\E_[P_{D|F}P_{D'|F}]|| P_{D} P_{D'}) \\ &\leq \min\{I(F;D), I(F;D')\},
\end{aligned}
\end{equation}
\end{claim}
where $KL$ denotes the Kullback–Leibler divergence.
A similar proof is given by \citep{clusterinfonce}, and we modify it to fit this paper in Appendix A.
We use $L_{NCE}$ to denote $L'_{NCE}$ for simplicity in the rest of our paper.

\subsection{Query-by-distribution}
\label{sec:query-by-distribution}
Another problem is that the amount of information contained in each data point is extremely small, which increases the number of query steps and makes the query network hard to optimize.
To alleviate this problem, we compress the number of query steps by querying a set of $m$ data points at each query step instead of only one single data point.

However, simply increasing the output dimension of the query network to get $m$ data points is sub-optimal 
because the number of data points needed in symbolic regression changes depending on the specific problem (\ie $m$ changes), which means that the output dimension should be changed accordingly.
Since a set can be presented by exponential family distributions~\cite{Sun2019InformationGeometricSE}, we represent the query result as a Normal distribution, from which we can flexibly sample any amount of latent vectors and use them to generate the final queries.
Also, the training process benefits from the diversity caused by sampling which can help it avoid the local minima.
We choose the Normal distribution due to its good properties: (1) It has a differentiable closed form KL divergence which is convenient for optimization. (2) The product of two Normal distributions (i.e. the intersection of two sets) is still a scaled Normal distribution~\cite{Huang2022NeuralPS}.
Details are shown in the next subsection.

\subsection{QUOSR}
Next, we will introduce the architecture and training details.
Architecture details are shown in Figure~\ref{fig:overview}.
As mentioned before, QUOSR adopts a siamese architecture where the two branches share parameters.
Each branch mainly contains two parts: a physical system $f$ which is used to be queried by generated $\vx$,
and a query network which receives historical data points $\{(\vx_i, y_i)\}_i$ and generates the next queries $\vx$.
The query network consists of an encoder that embeds the historical data points, and a decoder that receives the data embedding and generates the next queries.
Details are described in the following.
\paragraph{Encoder.}
The design of the encoder follows \citet{Ren2020BetaEF} and \citet{Huang2022NeuralPS}.
First, given a set of data points $D=\{(\vx_i, y_i)\}_{i=1}^{N}$, each data point is projected into a Normal distribution in latent space.
This distribution can be seen as the representation of the set of candidate expressions that satisfy the given data point: 
\begin{equation}
\label{eq:encoder IO}
[\vmu_i, log({\vsigma_i}^2)] = MLP_{data}((\vx_i, y_i)),
\end{equation}
where $[]$ denotes concatenation, and $\vmu_i$ and $\vsigma_i$ represents Normal distribution $\mathcal{N} (\vmu, \vsigma)$.
Then, for the $N$ distributions, we take them as independent ones (\ie input order irrelevant) and calculate their intersection as the final embedding of data points $D$:
\begin{equation}
\label{eq:intersection}
[\vmu, log(\vsigma^2)] = \sum_{i=1}^N w_i[\vmu_i, log({\vsigma_i}^2)]
,
\end{equation}
\begin{equation}
\label{eq:weight}
w_i = \frac{exp(MLP_{attention}([\vmu_i, log(\vsigma_i^2)]))}{\sum_{j=1}^N exp(MLP_{attention}(\vmu_j, log(\vsigma_j^2)))},
\end{equation}
where $MLP$ means multi-layer perceptron.
The further illustration of the intersection is beyond the scope of this paper and please refer to their original papers.

\paragraph{Decoder.}
The decoder consists of two parts: generating a Normal distribution and sampling $m$ points from it. 
Different from the distribution produced by the encoder which represents the set of candidate expressions, this distribution represents the set of possible queries as mentioned in the last subsection.
\begin{equation}
\label{eq:decoder set}
[\vmu_q, log({\vsigma_q}^2)] = MLP_{query}[\vmu, log(\vsigma^2)]
,
\end{equation}
where the subscript $q$ denotes "query".
Then, we sample $m$ points from the Normal distribution $\mathcal{N} (\vmu, \vsigma)$ and project them back to $\sR^M$ with an MLP.
\begin{equation}
\label{eq:converter}
\vx_i = MLP_{inversion}(\vv_i)
,
\end{equation}

\paragraph{Training.}
At the beginning of the query process, we sample two groups of data points uniformly as the initial condition of the query process (the number of initial data points equals the number of data points queried in each step).
Since $L_{NCE}$ is calculated by measuring the similarity between two different sets of data points $d$ and $d'$, we maintain two query processes with different initial conditions simultaneously.
Specifically, in each step of query, QUOSR receives $N$ historical data points $\{(\vx_i, y_i)\}_{i=1}^N$ as input, generates $m$ new data $\{\vx_i\}_{i=N+1}^{N+m}$, and then gets the corresponding $\{y_i\}_{i=N+1}^{N+m}$ from f. These $N+m$ data points $\{(\vx_i, y_i)\}_{i=1}^{N+m}$ are taken as the next step input of QUOSR, and this process repeats.
As mentioned in Equation~\ref{eq:intersection}, the data embedding represents Normal distributions, so the similarity of two sets of data points is measured with Kullback-Leibler divergence:
\begin{equation}
\label{eq:similarity}
sim(d,d') = KL(N(\vu_d, \vsigma_d)||N(\vu_d', \vsigma_d'))
\end{equation}
We merge d’ into d in Equation~\ref{eq:modified infonce} by crossing their elements, thus the loss function in Equation~\ref{eq:modified infonce} is finally rewritten as
\begin{equation}
\label{eq:infonce_simclr}
L_{NCE} = \frac{1}{2N} \sum_{i=1}^N [l(d_{2i-1}, d_{2i}) + l(d_{2i}, d_{2i-1})],
\end{equation}

\begin{equation}
\label{eq:infonce_simclr_l}
l(d_i,d_j) = -\log \frac{exp(sim(d_i,d_j)/\tau)}{ \sum_{k=1}^{2N} 1_{[k \neq i]} exp(sim(d_i,d_k)/\tau) }
\end{equation}
where $d_{2i-1}$ denotes $d$, $d_{2i}$ denotes $d'$ and $\tau$ denotes a temperature parameter.
See Figure~\ref{fig:overview} and Algorithm 2 in Appendix C for details.

\section{Experiments}
We study the ability of QUOSR to find informative points in experiments. First, we combine QUOSR with a symbolic regression method and present the main results. Then some cases are given to visualize the strategy learned by QUOSR. And at last, we do the ablation experiments.

\subsection{Settings}
We combine QUOSR with a transformer-based method named SymbolicGPT~\cite{Valipour2021SymbolicGPTAG} which achieves competent performance in symbolic regression.
SymbolicGPT uses PointNet~\cite{qi2017pointnet} to encode any number of data points and then generates the skeleton of an expression character by character using GPT where constants are represented by a $<C>$ token.
To learn these constants in the skeleton, SymbolicGPT employs BFGS optimization~\cite{fletcher2013practical}.

The training process is split into three stages: training QUOSR, generating a new dataset and finetuning SymbolicGPT. 
We train QUOSR using expressions in the dataset of~\citet{Valipour2021SymbolicGPTAG} which contains about 500k one-variable expressions.
The data points used for generating expressions are uniformly sampled from the range of $[-3.0,3.0]$.
The max query times $K$ is set to 9, thus we first uniformly sample 3 points and then generate the other 27 data points in the following 9 query steps. We limit QUOSR to generate values of $x \in [-3.0,3.0]$ to fit the range of the original dataset.
Then we generate a new dataset named $Query$ for all expressions.
For comparison, we also sample $x$ from $\mathcal{U}(-3.0, 3.0)$ and $\mathcal{N}(0,1)$ and generate another two datasets, $Uniform$ and $Normal$.
We finetune the pretrained model of SymbolicGPT on these three datasets.

The original test set of SymbolicGPT contains 966 expressions. For each expression, 30 data points sampled uniformly in the range of $[-3.0,3.0]$ are used to generate expressions, and another 30 data points sampled uniformly in the range of $[-5.0,-3.0] \cup [3.0,5.0]$ are used to evaluate the accuracy of the predicted expression.
We replace data points used for generating expressions with the ones sampled with three methods: QUOSR, uniform sampling, and normal sampling, and use original test data points to evaluate two metrics with three models   respectively: the proportion of  $\log MSE_N < -10$ and the average $R^2$~\cite{Kamienny2022EndtoendSR}:
\begin{equation}
\label{eq:logmse}
MSE_N = \sum_{i=1}^{N_{test}} \frac{(y_i-\hat{y}_i)^2}{||\vy + \epsilon||_2}
\end{equation}

\begin{equation}
\label{eq:r2}
R^2 = \max ( 0, 1-\frac{\sum_i^{N_{test}}(y_i-\hat{y}_i)^2}{\sum_i^{N_{test}}(y_i-\overline{y})^2} ),
\end{equation}
where $\epsilon$ is set to 0.00001 to avoid division by zero and $\overline{y} = \frac{1}{N_{test}}\sum_{i=0}^{N_{test}}y_i$. We set the lower bound of $R^2$ to 0 since a bad prediction can lead $R^2$ to be an extremely large negative number.

\subsection{Results}

\begin{figure}[ht]
    \centering
    \includegraphics[scale=0.70]{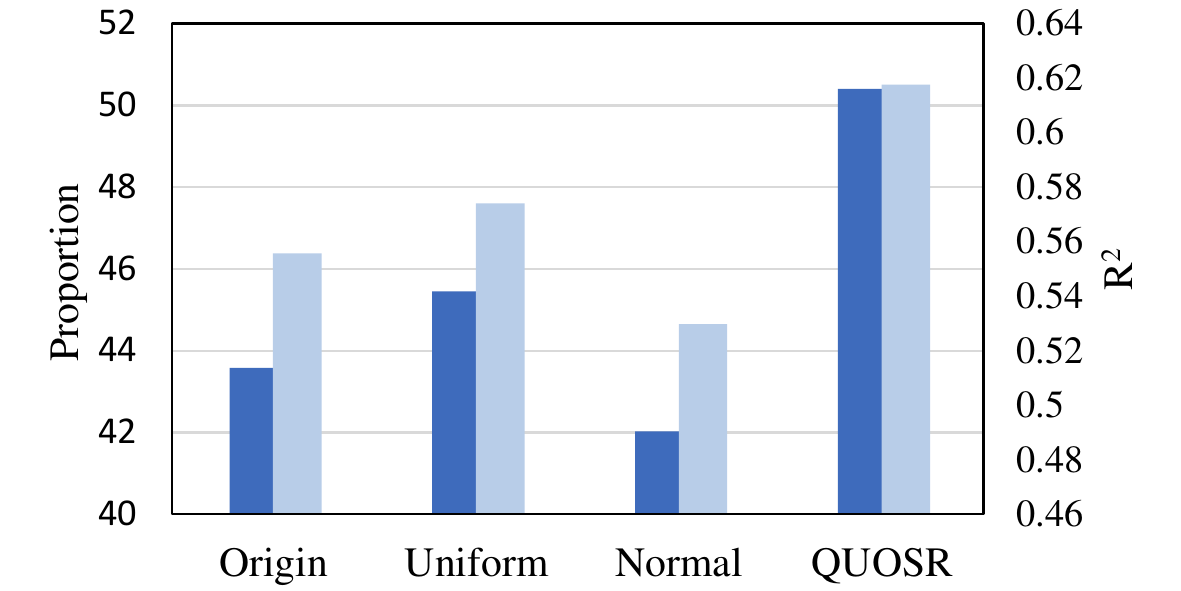}
    \caption{The performance of the origin SymbolicGPT model and another three models finetuned on $Uniform$, $Normal$ and $Query$.}
    \label{fig:symbolicgpt}
\end{figure}

Figure~\ref{fig:symbolicgpt} shows the proportion of predicted expressions whose $\log MSE_N < -10$ and the average of $R^2$.
Though results of uniform sampling improve after finetuning, QUOSR performs the best in both metrics, which indicates that
in the online setting of symbolic regression, QUOSR can explore more informative data points than uniform sampling and thus helps SymbolicGPT achieve a better result.

\begin{figure}[ht]
    \centering
    \includegraphics[scale=0.53]{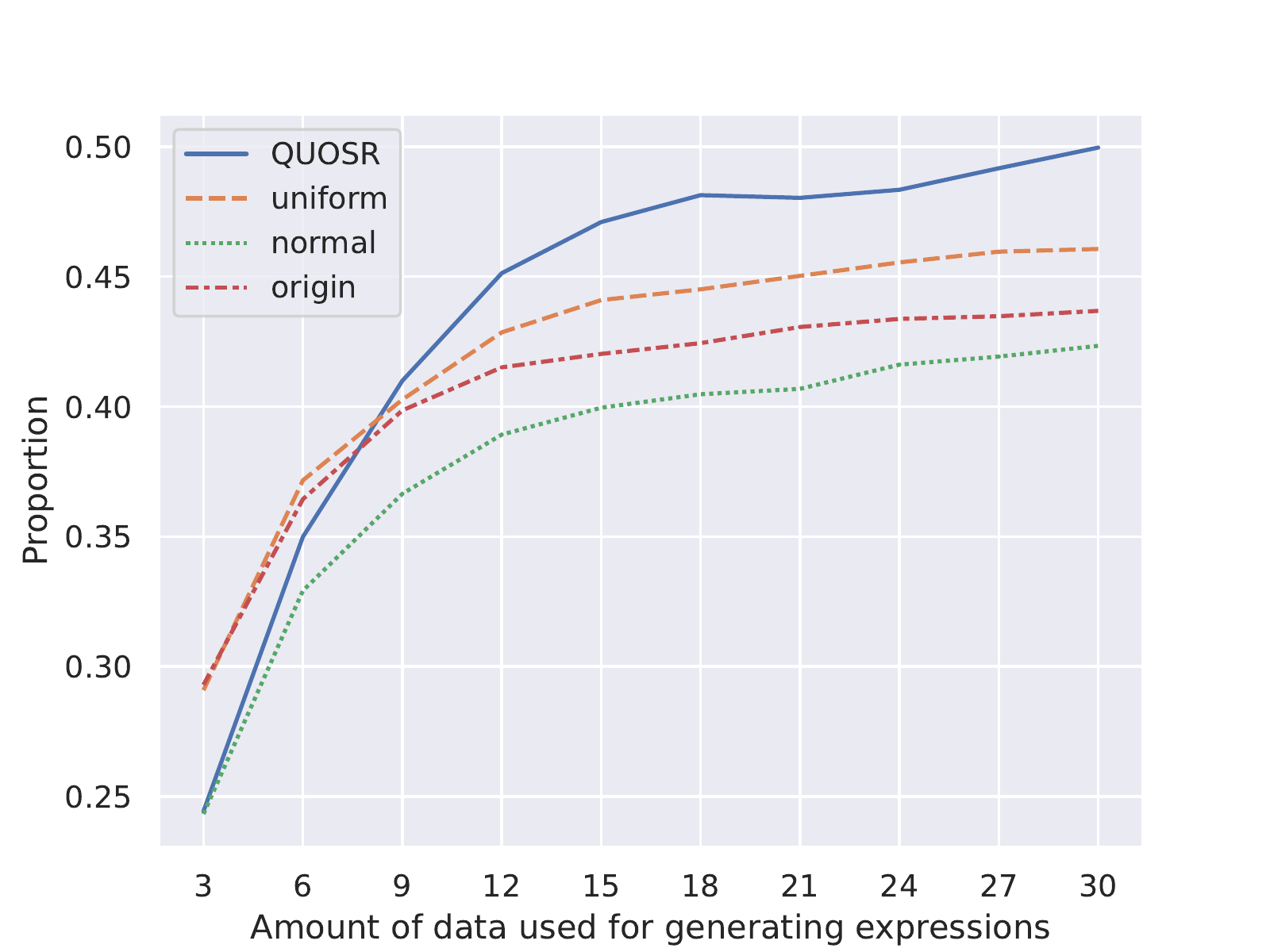}
    \caption{Results for different amount of data points.}
    \label{fig:query_steps}
\end{figure}

We also evaluate the performance of QUOSR by each query steps, shown in Figure~\ref{fig:query_steps}.
Since the more information a data point contains, the sooner it will be queried, the performance of QUOSR is significantly improved in the first four steps and grows steadily in the next five.

\begin{figure*}[ht]
    \centering
    \includegraphics[scale=0.41]{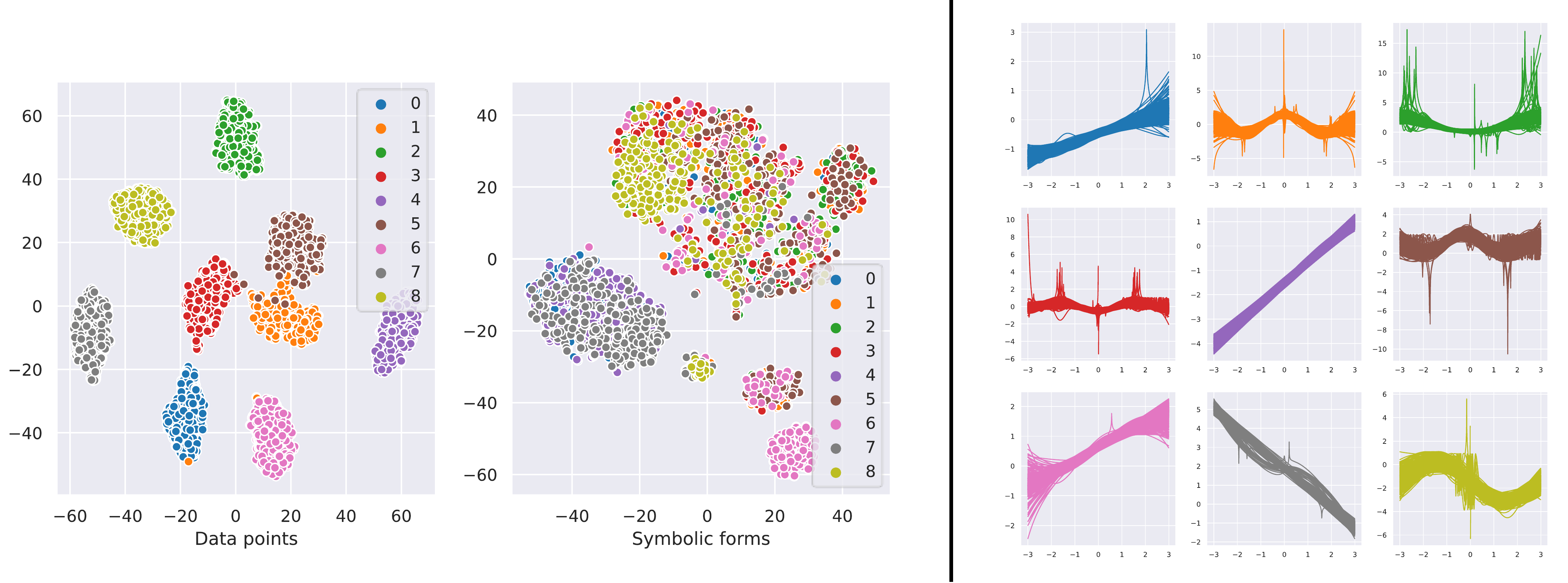}
    \caption{The clustering results of different embedding methods (Left and Mid). Different colors represent different sets of expressions that have similar functionality (Right).}
    \label{fig:embedding}
\end{figure*}

\begin{figure}[ht]
    \centering
    \includegraphics[scale=0.4]{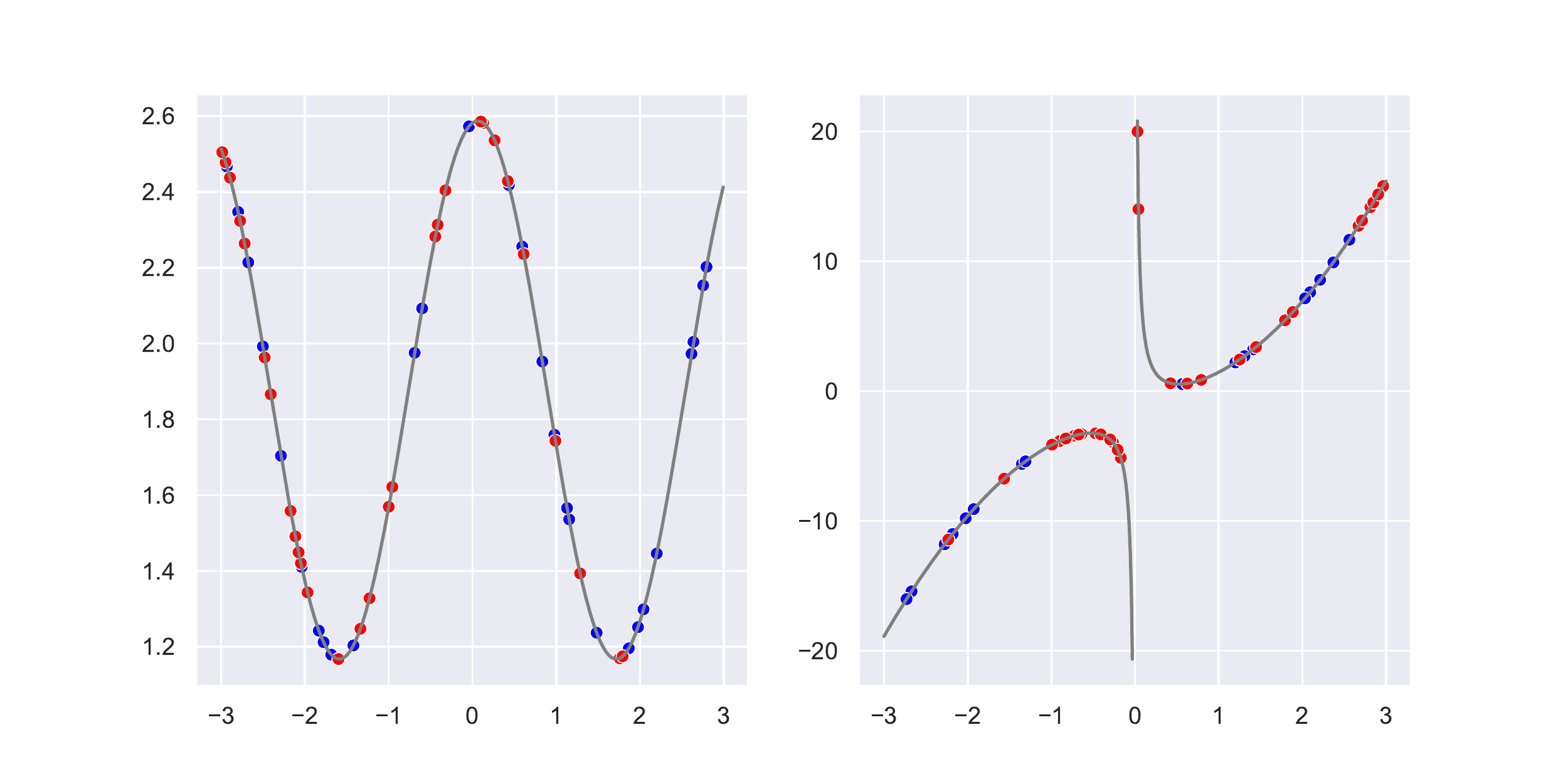}
    \caption{Graphs of two expressions in which QUOSR outperforms uniform sampling. The blue points are sampled uniformly and the red ones are generated by QUOSR.}
    \label{fig:cases}
\end{figure}

\subsection{Case Study}
Figure~\ref{fig:embedding} gives a visualization of the latent space using t-SNE~\cite{van2008visualizing}.
We cluster data points embeddings (Figure~\ref{fig:embedding} Left), symbolic forms embeddings (Figure~\ref{fig:embedding} Mid), and functions (into 9 classes, Figure~\ref{fig:embedding} Right).
To see which embedding method matches the functionality better, we label 9 classes in Figure~\ref{fig:embedding} Right with 9 different colors and color Figure~\ref{fig:embedding} Left and Mid with corresponding colors.
Graphs of expressions in the same cluster are close, but their symbolic form embeddings are chaotic.

Figure~\ref{fig:cases} presents
graphs of two expressions in which QUOSR outperforms uniform sampling. Data points sampled uniformly are marked with blue dots and the ones generated by QUOSR are marked in red. We observe two interesting phenomena:
(1) QUOSR pays more attention to the neighborhood of extreme points and discontinuity points.
(2) For periodic functions, QUOSR samples multiple points in one cycle, and only a few points in other cycles which might be used to judge their periodicity.

\subsection{Ablation Study}
Table~\ref{tab:ablation} presents results for the ablation study over the following four variants:

(1) Methods of intersection: \textbf{attention} in Equation~\ref{eq:intersection}, \textbf{mean} in Equation~\ref{eq:mean} or \textbf{max} in Equation~\ref{eq:max}:
\begin{equation}
\label{eq:mean}
[\vmu, log(\vsigma^2)] = [\frac{1}{n} \sum_{i=1}^N \vmu_i, \log(\frac{1}{n} {\sum_{i=1}^N \vsigma_i^2})]
,
\end{equation}

\begin{equation}
\label{eq:max}
[\vmu, log(\vsigma^2)] = [\max \vmu_i, \log({\max \vsigma_i^2})]
,
\end{equation}

(2) The similarity function: \textbf{KL} divergence in Equation~\ref{eq:similarity} or cosine similarity (\textbf{cos}) in Equation~\ref{eq:cosine}:

\begin{equation}
\label{eq:cosine}
sim(d,d') = \frac{[\vu_d, \vsigma_d] \cdot [\vu_d', \vsigma_d']^T}{\Vert [\vu_d, \vsigma_d]\Vert \Vert [\vu_d', \vsigma_d']\Vert }
,
\end{equation}

(3) Representation methods of physical system $f$: data points (\textbf{data}) or symbolic forms (\textbf{expr}).

(4) Query strategies: query-by-distribution (\textbf{QBD}), query-by-set (\textbf{QBS}) in Equation~\ref{eq:QBS} or query-by-point (\textbf{QBP}) Equation~\ref{eq:QBP}.

\begin{equation}
\label{eq:QBS}
[\{\vx_i\}_{i=1}^m] = MLP_{QBS}([\vmu, log(\vsigma^2)])
,
\end{equation}

\begin{equation}
\label{eq:QBP}
\vx_i = MLP_{QBP}([\vmu, log(\vsigma^2)])
,
\end{equation}

\begin{table}[t]
\centering
\begin{tabular}{llll|ll}
    Intersect & Sim & Rep & Stra & Proportion & $R^2$ \\
\hline
    
    mean & KL & data & QBD & 41.20 & 0.5370 \\
    max & KL & data & QBD  & 45.65 & 0.5774 \\
\hline
    attention & cos & data & QBD  & 41.72 & 0.5380 \\
\hline
    attention & KL & expr & QBD  & 44.20 & 0.5721 \\
    attention & KL & expr & QBS & 40.89 & 0.5290 \\
\hline
    attention & KL & data & QBS & 41.61 & 0.5388 \\
    attention & KL & data & QBP & N/A & N/A \\
\hline
    attention & KL & data & QBD & \textbf{50.41} & \textbf{0.6177}\\
\end{tabular}
\caption{Ablation over interaction methods (Intersect), the similarity function (Sim), expression representation (Rep) and query strategies (Stra). N/A means the out of memory error code occurs when training.}
\label{tab:ablation}
\end{table}

The results are presented in table~\ref{tab:ablation}. We can conclude that:
(1) Experiments with the \textbf{KL} similarity and \textbf{attention} intersection get a better result than others, suggesting the advantage of projecting data points to a Normal distribution
(2) The performance drops when using symbolic forms of expressions, which indicates that representing an expression with its data points is more appropriate than with its symbolic forms.
(3) Among all query strategies, our query-by-distribution strategy performs the best. 
The query-by-point strategy does not give out useful results due to its optimization difficulty.
The query-by-set strategy performs worse than ours probably because sampling from distribution results in more diversity which helps the query network avoid local minima in training.

\section{Conclusion}
In this work, we propose a query-based framework QUOSR to tackle online symbolic regression.
To maximize the mutual information between data points and expressions, we modify InfoNCE loss and thus a good expression encoder is unnecessary.
Furthermore, we use the query-by-distribution strategy instead of query-by-point, and thus the amount of information obtained in each query step increases.
Combining with modern offline symbolic regression methods, QUOSR achieves a better performance.
We take the query-based framework for multi-variable expressions as our future work for its challenge in spurious multi-variable expressions.

\section{Acknowledgments}
This work is partially supported by the NSF of China(under Grants 61925208, 62222214, 62102399, 62002338, U22A2028, U19B2019), Beijing Academy of Artificial Intelligence (BAAI), CAS Project for Young Scientists in Basic Research(YSBR-029), Youth Innovation Promotion Association CAS and Xplore Prize.

\bibliography{aaai23}

\clearpage

\appendix

\section{A \@ Theoretical Analysis}
\label{appendixA}

\subsection{Proof of Claim 1}

As the average number of query steps is equivalent to the average decision path, we intent to minimize the latter. This optimization problem has been solved by Huffman~\cite{Gallager1968InformationTA}.
\begin{equation}
\label{eq:huffman_bound}
\begin{aligned}
cH(F) \leq \bar{L}_{h} < cH(F) + 1,
\end{aligned}
\end{equation}
where $H(F)$ denotes the entropy of $F$, $\bar{L}_{h}$ denotes the minimum of the average decision path and $c$ is a constant.
Suppose several data points $D = \{\vx_i, y_i\}_{i=1}^m$ have been queried which means the length of the previous path $\bar{L}_{h1} = m$, the rest part $\bar{L}_{h2}$ can be bound with
\begin{equation}
\label{eq:rest_huffman_bound}
\begin{aligned}
cH(F|D) \leq \bar{L}_{h2} < cH(F|D) + 1.
\end{aligned}
\end{equation}
Thus the minimum length of average decision path after queries satisfies
\begin{equation}
\label{eq:query_huffman_bound}
\begin{aligned}
m + cH(F|D) \leq \bar{L}_{h1} + \bar{L}_{h2} < m + cH(F|D) + 1.
\end{aligned}
\end{equation}
The gap between ideal minimum length and minimum length after queries is
\begin{equation}
\label{eq:gap_bound}
\begin{aligned}
|\bar{L}_{h1} + \bar{L}_{h2} - \bar{L}_{h}| &< m + cH(F|D) + 1 - cH(F) \\
&= m + 1 -cI(F;D).
\end{aligned}
\end{equation}
Thus the optimal queries are
\begin{equation}
\label{eq:optimal_queries}
\begin{aligned}
D^* &= \arg\min_D |\bar{L}_{h1} + \bar{L}_{h2} - \bar{L}_{h}| \\
&= \arg\max_D I(F;D).
\end{aligned}
\end{equation}

\subsection{Proof of Claim 2}

We define $X = P_{D|F}P_{D'|F}$ and $Y = P_{D} P_{D'}$, thus according to~\citet{clusterinfonce, song2020multi}:
\begin{equation}
\begin{aligned}
\label{eq:infonce_1}
-L'_{NCE} &= \E_{(d_i, d'_i)\sim\E[P_{d|f}P_{d'|f}]}[log(\frac{exp(sim(d_i, d'_i))}{\frac{1}{N}\sum_{j=1}^N exp(sim(d_i, d'_j))})] \\
&= \E_{(d, d'_1)\sim X, (d, d'_{2:n}) \sim Y} [log(\frac{exp(sim(d, d'_1))}{\frac{1}{N}\sum_{j=1}^N exp(sim(d, d'_j))})] \\
&\leq KL(\E_X||Y) \\
&= KL(\E [P_{D|F}P_{D'|F}]|| P_{D} P_{D'})
\end{aligned}
\end{equation}

Then we give proof of 
\begin{equation}
\label{eq:kl_I1}
\begin{aligned}
KL(\E_[P_{D|F}P_{D'|F}]|| P_{D} P_{D'}) \leq \min\{I(F;D), I(F;D')\}.
\end{aligned}
\end{equation}

\begin{equation}
\label{eq:kl_I2}
\begin{aligned}
& I(F;D) - KL(\E_[P_{D|F}P_{D'|F}]|| P_{D} P_{D'}) \\
= & \int_F \int_D p(F,D)\log \frac{p(F,D)}{p(F)p(D)} dF dD - \\
& \int_F p(F) \int_D p(D|F) \int_{D'} p(D'|F) \\
& \log \frac{\int_{F'} p(F')p(D|F')p(D'|F')dF'}{p(D)p(D')} dD d{D'} dF \\
= & \int_F p(F) \int_D p(D|F) \log \frac{p(D|F)}{p(D)} dD dF - \\
& \int_F p(F) \int_D p(D|F) \int_{D'} p(D'|F) \\
& \log \frac{\int_{F'} p(F'|D')p(D|F')dF'}{p(D)} dD d{D'} dF \\
= & \int_F p(F) \int_D p(D|F) \int_{D'} \log \frac{p(D|F)}{\int_{F'} p(F'|D')p(D|F')dF'} dD d{D'} dF \\
\geq & - \int_F p(F) \int_D p(D|F) \int_{D'} \\
& (\frac{\int_{F'} p(F'|D')p(D|F')dF'}{p(D|F)} - 1) dD d{D'} dF \\
= & 0
\end{aligned}
\end{equation}
Similarly $KL(\E_[P_{D|F}P_{D'|F}]|| P_{D} P_{D'}) \leq I(F;D')$, thus
\begin{equation}
\label{eq:kl_I3}
\begin{aligned}
KL(\E_[P_{D|F}P_{D'|F}]|| P_{D} P_{D'}) \leq \min\{I(F;D), I(F;D')\}.
\end{aligned}
\end{equation}

\section{B \@ Experimental Details}
\label{appendixB}

We set the max query steps $K$ to 9. The dimension of $\vu$ and $\vsigma$ is set to 256. We use a three-layer transformer to encode symbolic forms of expressions whose hidden state is a 512 vector. For training, we use the SGD optimizer with a learning rate 1e-3. The batch size is set to 256.

\begin{figure}[th]
    \centering
    \includegraphics[scale=0.60]{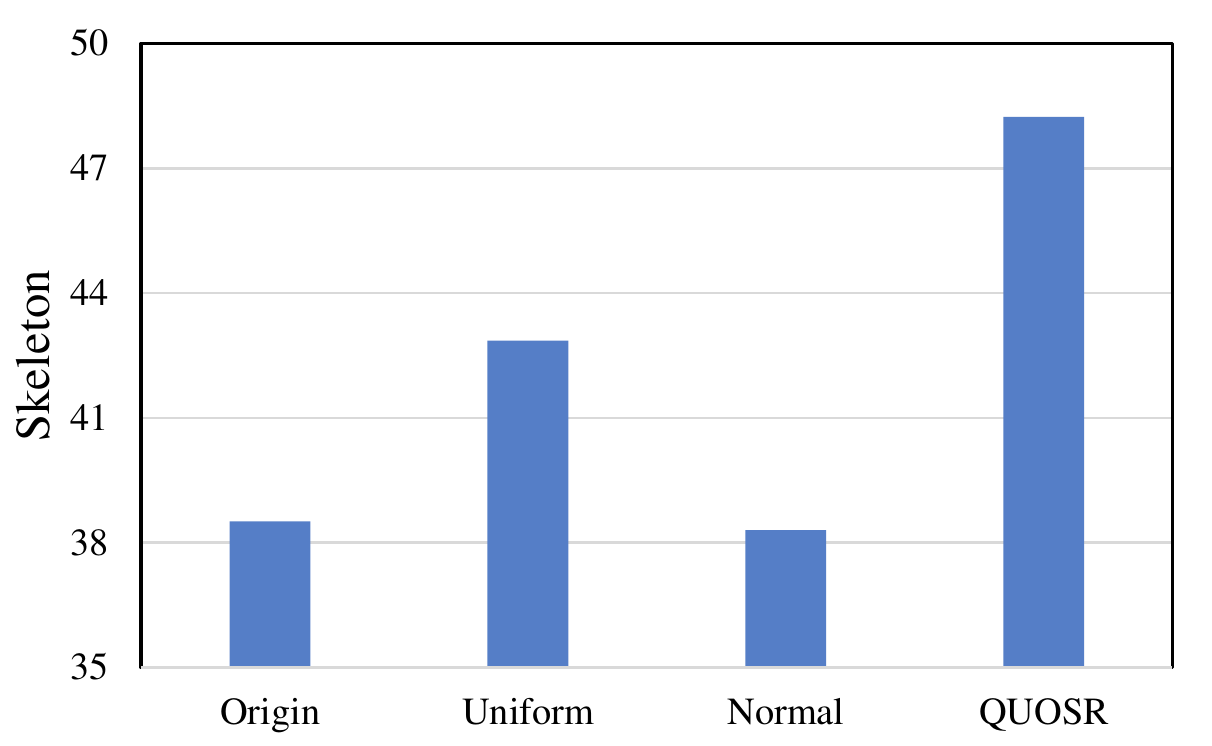}
    \caption{The performance of the origin SymbolicGPT model and another three models finetuned on $Uniform$, $Normal$ and $Query$.}
    \label{fig:skeleton}
\end{figure}

\begin{figure}[th]
    \centering
    \includegraphics[scale=0.50]{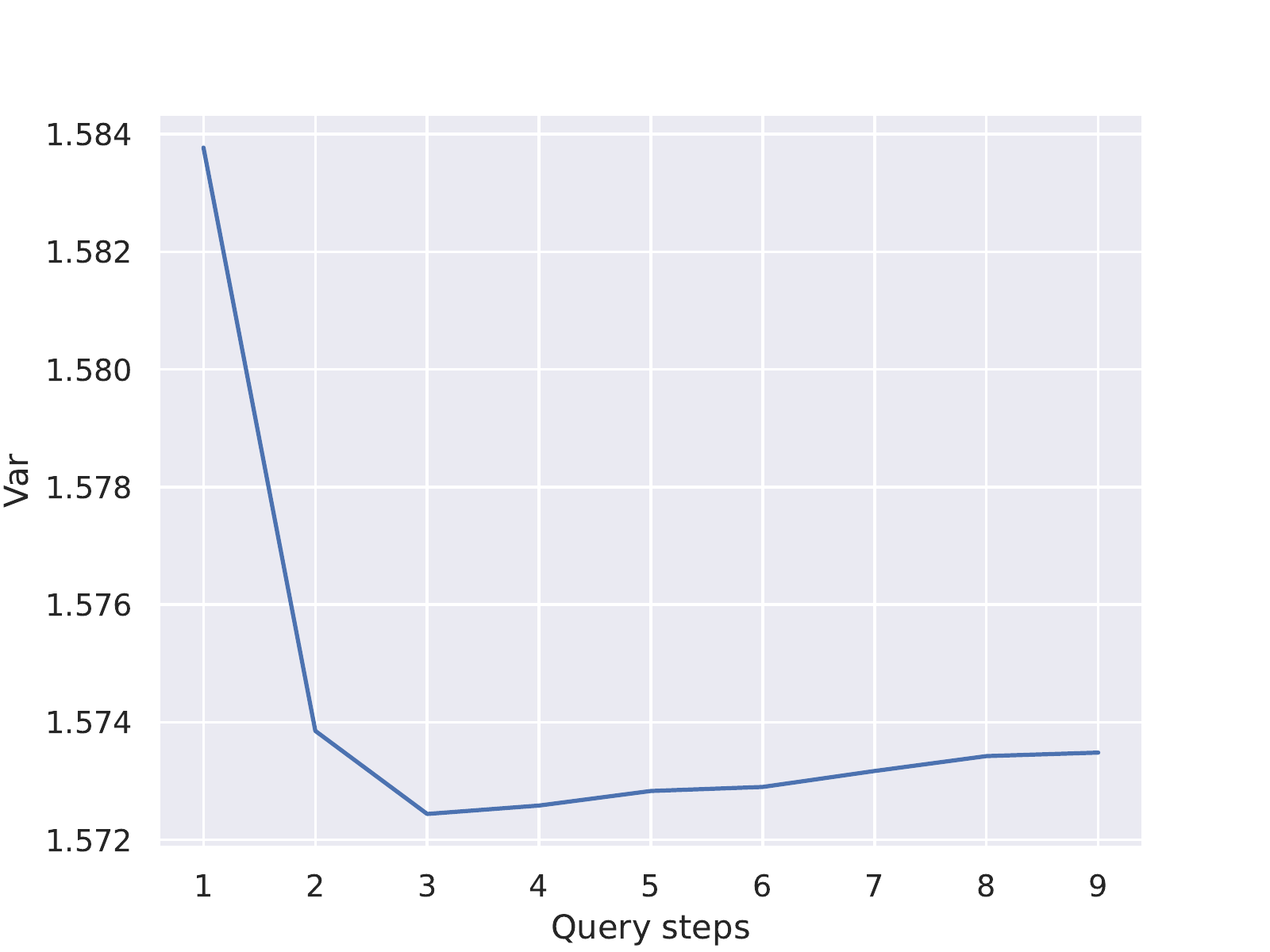}
    \caption{Variance of the Normal distribution in the latent space.}
    \label{fig:var_query_steps}
\end{figure}

\begin{table}[th]
\centering
\begin{tabular}{l|llll}
    Metric & Origin & Uniform & Normal & QUOSR \\
\hline
    
    Proportion & \textbf{67.35} & 44.90 & 61.22 & 63.27 \\
    $R^2$ & 0.80 & 0.75 & 0.80 & \textbf{0.86} \\
    Isclose & 0.76 & 0.60 & 0.72 & \textbf{0.79}

\end{tabular}
\caption{The performance of QUOSR combined with Nesymres.}
\label{tab:thatscales}
\end{table}

Since SymbolicGPT outputs the skeleton of expressions, we also measure skeleton match in the test set which means expressions with constant replaced with a $<C>$ token are correct. Skeleton match is a meaningful metric. It describes whether the SR method correctly judges the relationship between the variables, which is what experimenters care about.
Figure~\ref{fig:skeleton} shows the results. QUOSR obviously outperforms, probably because QUOSR samples data points at different intervals and focuses on extreme points.

Figure~\ref{fig:var_query_steps} shows the variance $log(\vsigma^2)$ of the Normal distributions in the latent space which represents the entropy of the Normal distribution.
As the amount of data points increases, the size of possible expression set becomes small, and the entropy of the normal distribution gradually decreases.
It becomes stable from step four, which leads to our performance improving at a slower rate.

We combine QUOSR with another SR method named Nesymres~\cite{Biggio2021NeuralSR} and present the results in table~\ref{tab:thatscales}. The test set contains 150 expressions. The metric $Isclose$ is the proportion of expressions for over 95\% of points of which $numpy.isclose(y, \hat(y))$ returns True. Results show that QUOSR is a general framework that can be aggregated to different SR approaches.



\section{C \@ Training Algorithm}
\label{appendixC}
Algorithm~\ref{alg:train} describes the training process of QUOSR.

\begin{algorithm}[h!]
    \caption{Training process}
    \label{alg:train}
    Train
    \begin{algorithmic}[1]
        \STATE Initialize max iterations $M$, max query times $K$ and sample amount $m$ for one query.
        \FOR{$ i \in \{1 \ldots M \}$}
            \STATE uniformly sample $d \gets \{ \vx_i, y_i\}^m$ and $d' \gets \{ \vx_j, y_j\}^m$ from the physical system
            \STATE $L \gets 0$
            \FOR{$ i \in \{1 \ldots K \}$}
                \STATE $\vmu, \vsigma \gets Encoder(d)$
                \STATE $\vmu', \vsigma' \gets Encoder(d')$
                \STATE $L \gets L + L_{NCE}([\vmu, \vsigma], [\vmu', \vsigma'])$
                \STATE $p \gets Decoder([\vmu, \vsigma])$
                \STATE $p' \gets Decoder([\vmu', \vsigma'])$
                \STATE $d \gets d \cup p$
                \STATE $d' \gets d' \cup p'$
            \ENDFOR
            \STATE Update parameters w.r.t. $L$
        \ENDFOR
    \end{algorithmic}
    Encoder($\{ \vx_i, y_i\}^N$)
    \begin{algorithmic}[1]
        \FOR{$ i \in \{1 \ldots N\}$}
            \STATE $[{\vmu}_i, {log(\vsigma_i^2)}] \gets MLP_{data}(\{\vx_i, y_i\})$ 
        \ENDFOR
        \FOR{$ i \in \{1 \ldots N\}$}
            \STATE $w_i \gets \frac{exp(MLP_{attention}([\vmu_i, log(\vsigma_i^2)]))}{\sum_{j=1}^N exp(MLP_{attention}(\vmu_j, log(\vsigma_j^2)))}$
        \ENDFOR
        \STATE $[\vmu, log(\vsigma^2)] = \sum_{i=1}^N w_i[\vmu_i, log(\vsigma_i^2)]$
        \RETURN $\vmu, log(\vsigma^2)$
    \end{algorithmic}
    Decoder($[\vmu, \vsigma]$)
    \begin{algorithmic}[1]
        \STATE $[\vmu_q, log(\vsigma_q^2)] \gets MLP_{query}([\vmu, log(\vsigma^2)])$
        \STATE sample $m$ vectors $\{ v_i\}_{i=0}^m$ from $N(\vmu_q, \vsigma_q)$
        \STATE $p \gets \emptyset$
        \FOR{$ i \in \{1 \ldots m\}$}
            \STATE $\vx_i = MLP_{inversion}(\vv_i)$
            \STATE interact with the physical system $f$, and get corresponding response $y_i = f(\vx_i)$
            \STATE $p \gets p \cup \{\vx_i, y_i\}$
        \ENDFOR
        \RETURN $p$
    \end{algorithmic}
\end{algorithm}

\end{document}